\documentclass{article}
\usepackage{spconf,amsmath}
\usepackage{pifont}
\usepackage{url}
\usepackage{soul}
\usepackage{epsfig}
\usepackage{amssymb}
\usepackage[normalem]{ulem}
\usepackage{multirow}
\usepackage{lipsum}  
\usepackage{xcolor}
\usepackage{soul}
\usepackage{amsfonts} 
\usepackage{booktabs}
\usepackage[ruled,vlined]{algorithm2e}
\usepackage[inline]{enumitem}
\usepackage{amsthm}
\usepackage{xcolor}
\usepackage{soul}
\usepackage{comment}
\usepackage{wrapfig}
\usepackage{booktabs}
\usepackage[symbol]{footmisc}
\usepackage{array}
\usepackage{algorithmic,algorithm2e,float}
\usepackage{xpatch}
\usepackage{hyperref}
\usepackage{graphicx}
\SetAlCapNameFnt{\scriptsize}
\SetAlCapFnt{\scriptsize}
\makeatletter
\xpatchcmd{\algorithmic}
  {\ALG@tlm\z@}{\leftmargin\z@\ALG@tlm\z@}
  {}{}
\makeatother
\makeatletter
\newcommand{\removelatexerror}{\let\@latex@error\@gobble}
\makeatother

\newcommand{\cmark}{\ding{51}}%
\newcommand{\xmark}{\ding{55}}%

\title{CLIP-TSA: CLIP-Assisted Temporal Self-Attention for\\ Weakly-Supervised Video Anomaly Detection}
\name{Hyekang Kevin Joo\textsuperscript{\dag}\thanks{\textsuperscript{\dag}Corresponding author's email address: hkjoo@cs.umd.edu},
Khoa Vo\textsuperscript{\ddag}, Kashu Yamazaki\textsuperscript{\ddag}, Ngan Le\textsuperscript{\ddag}}
\address{Dept. of Computer Science, University of Maryland, College Park, MD, USA\textsuperscript{\dag}\\Dept. of Computer Science and Computer Engineering, University of Arkansas, Fayetteville, AR, USA\textsuperscript{\ddag}}
\begin{document}
%
\maketitle
\begin{abstract}
Video anomaly detection (VAD) -- commonly formulated as a multiple-instance learning problem in a weakly-supervised manner due to its labor-intensive nature -- is a challenging problem in video surveillance where the frames of anomaly need to be localized in an untrimmed video. In this paper, we first propose to utilize the ViT-encoded visual features from CLIP, in contrast with the conventional C3D or I3D features in the domain, to efficiently extract discriminative representations in the novel technique. We then model temporal dependencies and nominate the snippets of interest by leveraging our proposed Temporal Self-Attention (TSA). The ablation study confirms the effectiveness of TSA and ViT feature. The extensive experiments show that our proposed CLIP-TSA outperforms the existing state-of-the-art (SOTA) methods by a large margin on three commonly-used benchmark datasets in the VAD problem (UCF-Crime, ShanghaiTech Campus and XD-Violence). Our source code is available at \href{https://github.com/joos2010kj/CLIP-TSA}{\url{https://github.com/joos2010kj/CLIP-TSA}}.
\end{abstract}

\section{Introduction}
\label{sec:intro}
Video action understanding is an active research field with many applications, e.g., action localization \cite{khoavo_aei_bmvc21, KhoaVo_Access, KhoaVo_ICASSP, he2022asm} action recognition \cite{vu2021teaching, vu20222+, sun2022human, quach2022non, vo2022contextual, xing2023svformer}, video captioning \cite{wang2020event,lei2020mart, yamazaki2022vlcap, yamazaki2023vltint}, etc \cite{hutchinson2020video}. Video anomaly detection (VAD) is the task of localizing anomalous events in a given video with three main paradigms, \textit{i.e.}, Fully-supervised (Sup.) \cite{liu2019exploring}
, Un-Sup \cite{gong2019memorizing}
, and Weakly-Sup \cite{ucf}. While it generally yields high performance, Fully-Sup VAD requires fine-grained anomaly labels (\textit{i.e.,} \textit{frame-level} normal/abnormal annotations), and the problem has traditionally suffered from the laborious nature of data annotation. In Un-Sup VAD, one-class classification (OCC)~\cite{claws} is a common approach, where the model is trained on only normal class samples with the assumption that unseen abnormal videos have high reconstruction errors. However, the performance of Un-Sup VAD is usually poor as it lacks prior knowledge of abnormality and from its inability to capture all normality variations \cite{chandola2009anomaly}. Compared to both Un-Sup and Sup VAD, Weakly-Sup VAD is considered the most practical approach because of its competitive performance and annotation efficiency by employing \textit{video-level} labels to reduce the cost of manual fine-grained annotations \cite{gcn}.

\begin{figure}[t]
\centering
  \includegraphics[width=\linewidth]{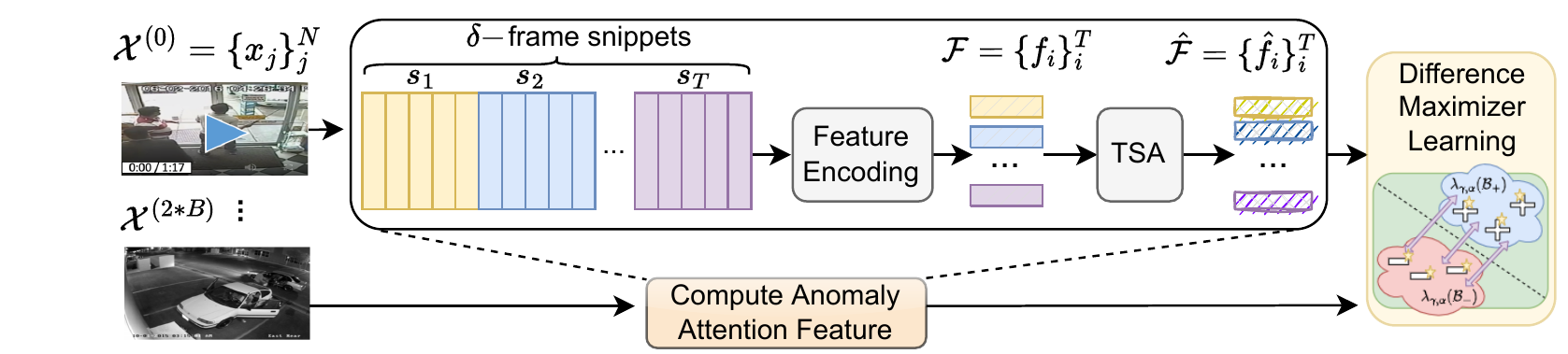}
  \caption{Overall chart of CLIP-TSA in train time, with $\mathcal{X}^{2*B}$ as input. Each video $\mathcal{X}$ consisting of $N$ frames (\textit{i.e.,} $\mathcal{X} = \{x_j\}_j^N$) is divided into a set of $\delta$-frame snippets $\{s_i\}_i^{{T}}$. Each $\delta$-frame snippet $s_i$ is represented by a V-L feature $f_i \in \mathbb{R}^d$. Then, the video is represented by $\mathcal{F} = \{f_i\}_i^{{T}}$. Our TSA is then applied to obtain anomaly attention feature $\mathcal{\hat{F}} = \{\hat{f}_i\}_i^T$. The features $\mathcal{\hat{F}}$ from $2*B$ videos then undergo difference maximization trainer to weakly-train anomaly classifier. \vspace{-5mm}}
\label{fig:overall} 
\end{figure}

In the Weakly-Sup VAD, there exist two fundamental problems. First, anomalous-labeled frames tend to be dominated by normal-labeled frames, as the videos are untrimmed and there is no strict length requirement for the anomalies in the video. Second, the anomaly may not necessarily stand out against normality. As a result, it occasionally becomes challenging to localize anomaly frames. Thus, the problem is commonly designed with multiple instance learning (MIL) framework~\cite{ucf}, where a video is treated as a bag containing multiple instances, each instance being a video snippet. The video is labeled as anomalous if any of its snippets are anomalous, and normal if all of its snippets are normal. Anomalous-labeled videos belong to the positive bag and normal-labeled videos belong to the negative bag. However, the existing MIL-based Weakly-Sup VAD approaches are limited in dealing with an arbitrary number of abnormal snippets in an abnormal video. To address such an issue, we are inspired by the differentiable top-K operator~\cite{topk} and introduce a novel technique, termed top-$\kappa$ function, that localizes $\kappa$ snippets of interest in the video with differentiable hard attention in the similar MIL setting to demonstrate its effectiveness and applicability to the traditional, popular setting. Furthermore, we introduce the Temporal Self-Attention (TSA) Mechanism, which generates the reweighed attention feature by measuring the degree of abnormality of snippets.



In addition, the existing approaches encode visual content by applying a backbone, \textit{e.g.}, C3D \cite{C3D}, I3D \cite{i3d_2017}, which are pre-trained on action recognition tasks. Different from the action recognition problem, VAD depends on discriminative representations that clearly represent the events in a scene. Thus, those existing backbones are not suitable because of the domain gap \cite{liu2019exploring}. To address such limitation, we leverage the success of the recent ``vision-language" (V-L) works \cite{Patashnik2021styleclip}, which have proved the effectiveness of feature representation learned via Contrastive Language-Image Pre-training (CLIP) \cite{radford2021learning, yamazaki2022vlcap, yamazaki2023vltint}. Our proposed CLIP-TSA follows the MIL framework and consists of three components: (i) Feature Encoding by CLIP; (ii) Modeling snippet coherency in the temporal dimension with our TSA; and (iii) Weakly-training the model with Difference Maximization Trainer.  
\section{Proposed Method}
\subsection{Problem Setup}
\label{sec:probsetup}




Let there be a set of weakly-labeled training videos $S = \{\mathcal{X}^{(k)}, y^{(k)}\}_{k=1}^{|S|}$, where a video $\mathcal{X}^{(k)} \in \mathbb{R}^{N_k \times W \times H}$ is a sequence of $N_k$ frames that are $W$ pixels wide and $H$ pixels high, and $y^{(k)} = \{0,1\}$ is the \textit{video-level} label of video $\mathcal{X}^{(k)}$ in terms of anomaly (\textit{i.e.,} 1 if the video contains anomaly). 
Given a video $\mathcal{X}^{(k)} = \{x_j\}|_{j=1}^{N_k}$, we first divide $\mathcal{X}^{(k)}$ into a set of $\delta$-frame snippets $\{s_i\}_{i=1}^{\bigr\lceil \frac{N_k}{\delta} \bigr\rceil}$. Feature representation of each snippet is extracted by applying a V-L model onto the middle frame. In this work, CLIP is chosen as a V-L model; however, it can be substituted by any V-L model. Thus, each $\delta$-frame snippet $s_i$ is represented by a V-L feature $f_i \in \mathbb{R}^d$ and the video $\mathcal{X}^{(k)}$ is represented by a set of video feature vectors $\mathcal{F}_k = \{f_i\}|_{i=1}^{T_k}$, where $\mathcal{F}_k \in \mathbb{R}^{T_k\times d}$ and $T_k$ is the number snippets of $\mathcal{X}^{(k)}$. 
To allow batch-training, the input features $\mathcal{F}$ come post-normalized into the uniform video feature length in its temporal dimension $T$ as follows~\cite{ucf}, with $\lfloor g \rfloor = \lfloor \frac{{T}_1}{T} \rfloor$ and $\lfloor g \rfloor = \lfloor \frac{{T}_2}{T} \rfloor$ for $\mathcal{F}_1$ and $\mathcal{F}_2$, respectively:
\begin{equation}
\begin{split}
    \mathcal{F} = \{f_{i'}\}|_{i'=1}^{T} = \frac{1}{\lfloor g \rfloor} \sum_{i=\lfloor g \times (i'-1) \rfloor}^{\lfloor g \times i' \rfloor} f_i
\end{split}
\label{eq:reshape}
\end{equation}


\subsection{Feature Encoding}
\label{sec:featureEnc}

We choose the middle frame $I_i$ that represents each snippet $s_i$. We first encode frame $I_i$ with the pre-trained Vision Transformer \cite{alexey} to extract visual feature $I^f_i$. We then project feature $I^f_i$ onto the visual projection matrix $L$, which was pre-trained by CLIP to obtain the image embedding $f_i =  L\cdot I^f_i$. Thus, the embedding feature $\mathcal{F}_k$ of video $\mathcal{X}$, which consists of ${T_k}$ snippets $ \mathcal{X} = \{s_i\}|_{i=1}^{T_k}$, is defined in Eq.~\ref{eq:fea_enc}b. Finally, we apply the video normalization as in Eq.~\ref{eq:reshape} onto the embedding feature to obtain the final embedding feature $\mathcal{F}$ as in Eq.~\ref{eq:fea_enc}c.
\begin{subequations}
\begin{eqnarray}
    f_i & = & L\cdot I^f_i \hspace{14mm}\text{where } f_i \in \mathbb{R}^d \\
    \mathcal{F}_k & = & \{f_i\}|_{i=1}^{T_k} \hspace{10.5mm}\text{where } \mathcal{F}_k \in \mathbb{R}^{T_k \times d} \\
    \mathcal{F} & = & Norm(\mathcal{F}_k) \hspace{6mm}\text{where } \mathcal{F} \in \mathbb{R}^{T \times d}
\end{eqnarray}
\label{eq:fea_enc}
\end{subequations}
\noindent
\begin{minipage}{0.25\textwidth}
\removelatexerror
\begin{algorithm}[H]
  \algsetup{linenosize=\tiny}
  \scriptsize
\DontPrintSemicolon
\SetNoFillComment
\hspace{-2.5mm}\KwData{Feature $\mathcal{F} \in \mathbb{R}^{T\times d}$, \\\hspace{3.5mm} Top snippet count $\kappa \in \mathbb{R}^1$}
\hspace{-2.5mm}\KwResult{Anomaly attent. feature $\hat{\mathcal{F}}$}

\hspace{-2.5mm}$\omega \gets \phi_s(\mathcal{F})$ 
\\
\hspace{-2.5mm}$\hat{\mathcal{V}} \gets \text{Top-}\kappa\text{ Score}(M, \kappa, \omega)$  
\\
\hspace{-2.5mm}$\tilde{\mathcal{V}} \gets $ Make $d$ clones of $\hat{\mathcal{V}}$
\\
\hspace{-2.5mm}$\tilde{\mathcal{F}} \gets $ Make $\kappa$ clones of $\mathcal{F}$ 
\\
\hspace{-2.5mm}$\mathcal{Q} \gets \tilde{\mathcal{V}} \otimes \tilde{\mathcal{F}}$
\\
\hspace{-2.5mm}$\hat{\mathcal{F}} \gets $ summation of $ \mathcal{Q} $ across dim $\kappa$ 
\\
\hspace{-2.5mm}return $\hat{\mathcal{F}}$ 
\caption{\\TSA mechanism $\sigma$ to produce anomaly attention features $\hat{\mathcal{F}}$}
\label{algo:tsa}
\end{algorithm}
\end{minipage}
\hfill
\begin{minipage}{0.24\textwidth}
\removelatexerror
\begin{algorithm}[H]
  \algsetup{linenosize=\tiny}
  \scriptsize
    \DontPrintSemicolon
\SetNoFillComment
\hspace{-2.5mm}\KwData{Sample count $M$, \\\hspace{4.2mm}Top snippet count $\kappa$, \\\hspace{4.2mm}Score vector $\omega$}
\hspace{-2.5mm}\KwResult{A stack of soft one-hot vectors $\hat{\mathcal{V}}$}
\hspace{-2.5mm}set $\bar{\omega}$ to $M$ clones of $\omega$ 
\\
\hspace{-2.5mm}set $\mathcal{G}$ to Gaussian noise 
\\
\hspace{-2.5mm}$\hat{\omega} \gets$ $\mathcal{G} \oplus \bar{\omega}$ 
\\
\hspace{-2.5mm}$\mathcal{U} \gets$ indices of top-$\kappa$ scores \\\hspace{7mm} across dim $M$ in $\hat{\omega}$ 
\\
\hspace{-2.5mm}$\mathcal{V} \gets$ one-hot encode $\kappa$ in $\mathcal{U}$ 
\\
\hspace{-2.5mm}$\hat{\mathcal{V}} \gets$ average of $\mathcal{V}$ across dim $M$ 
\\
\hspace{-2.5mm}return $\hat{\mathcal{V}}$ 
\caption{\\$\text{Top-}\kappa\text{ Score}$ function}
\label{algo:topk}
\end{algorithm}
\end{minipage}

\subsection{Temporal Self-Attention (TSA)}
\label{sec:tsa}
\begin{figure}[t]
\centering
  \includegraphics[width=\linewidth]{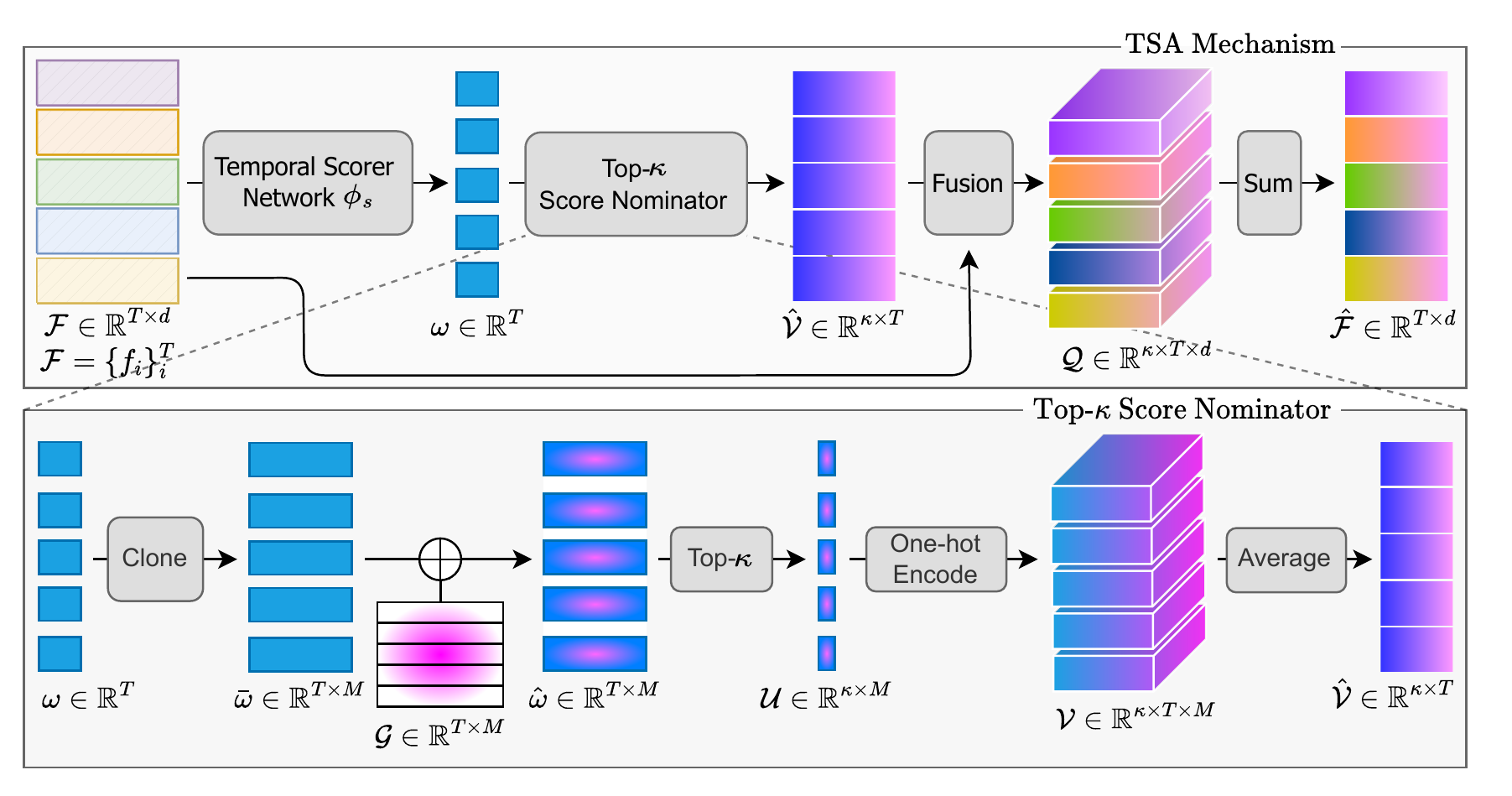}
  \caption{Top: Our proposed TSA mechanism to model the coherency between snippets.
  Bottom: Details of top-$\kappa$ score nominator network, which takes the score vector $\omega$ as its input and outputs a stack of soft one-hot vectors $\hat{\mathcal{V}}$.
  }
  \label{fig:tsa}
  \vspace{-5mm}
\end{figure}

Our proposed TSA mechanism aims to model the coherency between snippets of a video and select the top-$\kappa$ most relevant snippets. It contains three modules, \textit{i.e.}, (i) temporal scorer network, (ii) top-$\kappa$ score nominator, and (iii) fusion network, as shown in Fig.~\ref{fig:tsa} and Alg.~\ref{algo:tsa}. 
In TSA, the vision language feature $\mathcal{F} \in \mathbb{R}^{T\times d}$ (from \ref{sec:featureEnc} Feature Encoding) is first converted into a score vector $\omega \in \mathbb{R}^{T \times 1}$ through a \emph{temporal scorer network} $\phi_s$, \textit{i.e.}, $\omega = \phi_s(\mathcal{F})$. This network is meant to be shallow; thus, we choose a multi-layer perceptron (MLP) of 3 layers in this paper. The scores, each of which is representing the snippet $s_i$, are then passed into the \emph{top-$\kappa$ score nominator} to extract the $\kappa$ most relevant snippets from the video. The top-$\kappa$ score nominator is implemented by the following two steps. First, the scores $\omega\in \mathbb{R}^{T \times 1}$ are cloned $M$ times and the cloned score $\bar{\omega}\in \mathbb{R}^{T \times M}$ is obtained; $M$ represents the number of independent samples of score vector $\omega$ to generate for the empirical mean, which is to be used later for computing the expectation with noise-perturbed features. Second, Gaussian noise $\mathcal{G} \in \mathbb{R}^{T \times M}$ is applied to the stack of $M$ clones by Eq.~\ref{eq:gau} to produce $\hat{\omega} \in \mathbb{R}^{T \times M}$:
\begin{equation}
    \hat{\omega} = \mathcal{G} \oplus \bar{\omega} \qquad\text{where $\oplus$ is an element-wise addition}
\label{eq:gau}
\end{equation}

From the Gaussian-perturbed scores $\hat{\omega} \in \mathbb{R}^{T \times M}$, the indices of top-$\kappa$ snippets are selected based on the score magnitude independently across its $M$ dimension to represent the most relevant snippets and are later one-hot encoded into a matrix $\mathcal{V} = \{V_{i}\}_{i=1}^M$, with each $V_{i} \in \mathbb{R}^{\kappa \times T}$ containing a set of one-hot vectors. More specifically, we guide the network to place the attention on $\kappa$ magnitudes with the highest values because the Difference Maximization Trainer (Sec.~\ref{sec:dmt}) trains the anomalous snippets to have a high value and the normal snippets to have a low value. The matrix $\mathcal{V}$ is then averaged across its $M$ dimension to produce a stack of soft one-hot vectors $\hat{\mathcal{V}} \in \mathbb{R}^{\kappa \times T}$. Through the soft one-hot encoding mechanism, the higher amount of attention, or weight, is placed near and at the indices of top-$\kappa$ scores (\textit{e.g.}, [0, 0, 1, 0] $\rightarrow$ [0, 0.03, 0.95, 0.02]). Refer to Alg.~\ref{algo:topk} for its overview. 

Afterward, the stack of perturbed soft one-hot vectors $\mathcal{\hat{V}} \in \mathbb{R}^{\kappa \times T}$ is transformed into $\mathcal{\tilde{V}} \in \mathbb{R}^{\kappa \times T \times d}$ by making $d$ clones of $\hat{\mathcal{V}}$, and the set of input feature vectors $\mathcal{F} \in \mathbb{R}^{T \times d}$ is transformed into $\mathcal{\tilde{F}} \in \mathbb{R}^{\kappa \times T \times d}$ by making $\kappa$ clones of $\mathcal{F}$. Next, the matrices $\mathcal{\tilde{V}}$ and $\mathcal{\tilde{F}}$, which carry the reweighed information of snippets and represent the input video features, respectively, are fused together to create a perturbed feature $\mathcal{Q} \in \mathbb{R}^{\kappa \times T \times d}$ that represents the reweighed feature magnitudes of snippets based on previous computations as $\mathcal{Q} = \tilde{\mathcal{V}} \otimes \tilde{\mathcal{F}}$, where $\otimes$ is element-wise multiplication.

Then, each stack of perturbed feature vectors $Q \in \mathbb{R}^{\kappa \times d}$ within the perturbed feature $\mathcal{Q} = \{Q_{i}\}_{i=1}^T$ is independently summed up across its dimension $\kappa$ to combine the magnitude information of $Q_{i}$ into one vector $\hat{f_i} \in \mathbb{R}^{d}$. The reweighed feature vector, $\hat{f_i} \in \mathbb{R}^{d}$, which collectively forms $\hat{\mathcal{F}} = \{\hat{f_i}\}|_{i=1}^T$, is collectively obtained as output from TSA $\sigma$ to represent an anomaly attention feature $\hat{\mathcal{F}} \in \mathbb{R}^{T \times d}$. 


\subsection{Difference Maximization Trainer Learning}
\label{sec:dmt}


Given a mini-batch of $2*B$ videos $\{\mathcal{X}^{(k)}\}|_{k=1}^{2*B}$, each video $\mathcal{X}^{(k)}$ is represented by $\mathcal{F}_k = \{f_i\}|_{i=1}^T$ obtained by TSA (Section \ref{sec:tsa}). Let the input mini-batch be represented by $\mathcal{Z} = \{\mathcal{F}_k\}|_k^{2*B} \in \mathbb{R}^{2*B \times T \times d}$, where $B$, $T$, and $d$ denote the user-input batch size, normalized time snippet count, and feature dimension, respectively. The actual batch size is dependent on the user-input batch size, following the equation of $2*B$, because the first half, $\mathcal{Z}_{-} \in \mathbb{R}^{B \times T \times d}$, is loaded with a set of normal bags, and the second half, $\mathcal{Z}_{+} \in \mathbb{R}^{B \times T \times d}$, is loaded with a set of abnormal bags in order within the mini-batch. 
    
After the mini-batch undergoes the phase of TSA, it outputs a set of reweighed normal attention features $\hat{\mathcal{Z}}_{-} = \{\hat{\mathcal{F}_k}\}|_{k=1}^{B}$ and a set of reweighed anomaly attention features $\hat{\mathcal{Z}}_{+} = \{\hat{\mathcal{F}_k}\}|_{k=B}^{2*B}$. The reweighed attention features $\hat{\mathcal{Z}}$ are then passed into a convolutional network module $J$ composed of dilated convolutions~\cite{YuDilated} and non-local block~\cite{Wang_2018_CVPR} to model the long- and short-term relationship between snippets based on the reweighed magnitudes.
The resulting stack of convoluted attention features $\check{\mathcal{Z}} = \{\check{\mathcal{F}_k}\}|_k^{2*B}$, where $\check{\mathcal{Z}} \in \mathbb{R}^{2*B \times T \times d}$, is then passed into a shallow MLP-based score classifier network $C$ that converts the features into a set of scores $U \in \mathbb{R}^{2*B \times T \times 1}$ to determine the binary anomaly state of feature snippets. The set of scores $U$ is saved as part of a group of returned variables, for use in loss.

Next, each convoluted attention feature $\{\check{\mathcal{F}_k}\}|_{k=1}^{2*B}$ of the batch $\check{\mathcal{Z}}$ undergoes Difference Maximization Trainer (DMT). Leveraging the top-$\alpha$ instance separation idea employed by \cite{Li_2015_CVPR, ucf}, we use DMT, represented by $\upsilon_{\gamma, \alpha}$, in this problem to maximize the separation, or difference, between top instances of two contrasting bags, $\check{\mathcal{Z}}_{-}$ and $\check{\mathcal{Z}}_{+}$, by first picking out the top-$\alpha$ snippets from each convoluted attention feature $\check{\mathcal{F}_k}$ based on the feature magnitude. This produces a top-$\alpha$ subset $\dot{\mathcal{F}_k} \in \mathbb{R}^{\alpha \times d}$ for each convoluted attention feature $\check{\mathcal{F}_k} \in \mathbb{R}^{T \times d}$. Second, $\dot{\mathcal{F}_k}$ is averaged out across top-$\alpha$ snippets to create one feature vector $\ddot{\mathcal{F}_k} \in \mathbb{R}^{d}$ that represents the bag. The procedure is explained by Eq.~\ref{eq:lambda} below:
\vspace{-0.15cm}
\begin{equation}
    \lambda_{\gamma, \alpha}(\check{\mathcal{F}}) = \ddot{\mathcal{F}} = \max_{\Omega_{\alpha}(\check{\mathcal{F}}) \subseteq \{\check{f}_i\}_{i=1}^T} \frac{1}{\alpha} \sum_{\check{f}_i \in \Omega(\check{\mathcal{F}})} \check{f}_i 
\label{eq:lambda}
\end{equation}
\vspace{-0.15cm}

Here, $\lambda$ is parameterized by $\gamma$, which denotes its dependency on the ability of the convolutional network module $J$ (\textit{i.e.,} representation of $\ddot{\mathcal{F}}$ depends on the top-$\alpha$ positive instances selected with respect to $J$). Next, $\alpha$ denotes the size of $\Omega$, which represents a subset of $\alpha$ snippets from $\check{\mathcal{F}}$. Each representative vector $\ddot{\mathcal{F}}$ is then normalized to produce $\dddot{F} \in \mathbb{R}^{1}$. 
\begin{equation}
    \upsilon_{\gamma, \alpha}(\check{\mathcal{F}}_+, \check{\mathcal{F}}_-) = ||\lambda_{\gamma, \alpha}(\check{\mathcal{F}}_+)|| - ||\lambda_{\gamma, \alpha}(\check{\mathcal{F}}_-)||
\label{eq:separability}
\end{equation}
The separability is computed as in Eq.~\ref{eq:separability}, where $\check{\mathcal{F}}_- = \{\check{f}_{-,i}\}|_i^T$ represents a negative bag and $\check{\mathcal{F}}_+ = \{\check{f}_{+,i}\}|_i^T$ represents a positive bag. More specifically, we leverage the theorem of expected separability~\cite{Li_2015_CVPR} to maximize the separability of the top-$\alpha$ instances (feature snippets) from each contrasting bag: Under the assumption that $Z_+$ has $\epsilon \in [1, T]$ abnormal samples and ($T - \epsilon$) normal samples, $Z_{-}$ has T normal samples, and $T=|Z_+|=|Z_-|$, the separability of top-$\alpha$ instances (feature snippets) from positive (abnormal) and negative (normal) bags is expected to be maximized as long as $\alpha \leq \epsilon$. 
Afterward, to compute the loss, a batch of normalized representative features $\{\dddot{\mathcal{F}}_{normal}\}|_{k=1}^{B}$ and $\{\dddot{\mathcal{F}}_{abnormal}\}|_{k=B}^{2*B}$ are then measured for margins between each other. A batch of margins is then averaged out and used as part of the net loss together with the score-based binary cross-entropy loss computed using the score set $U$.

\begin{table}[tb]
    \centering
    \caption{Summarization of Three Datasets}
    \resizebox{0.9\linewidth}{!}{
    \begin{tabular}{c | c | c | cc | cc }
        \multirow{2}{*}{\textbf{Dataset}} &
        \multirow{2}{*}{\textbf{\# videos}} &
        \multirow{2}{*}{\textbf{Duration}} &
        \multicolumn{2}{c|}{\textbf{Train}} &
        \multicolumn{2}{c}{\textbf{Test}}
        \\
        \cline{4-7}
        &&&
        \textbf{Anomaly} & 
        \textbf{Normal} &
        \textbf{Anomaly} & 
        \textbf{Normal} \\
        \toprule
        UCF-Crime & 1,900 & 128 hours & 810 & 800 & 140 & 150 \\
        ShanghaiTech & 437 & 317,398 frames & 63 & 175 & 44 & 155 \\
        XD-Violence & 4,754 & 217 hours & 1,905 & 2,049 & 500 & 300 \\
        \bottomrule
    \end{tabular}
    }
    \label{tb:datasets}
    \vspace{-3mm}
\end{table}

\begin{table}[tb]
    \centering
    \caption{\vspace{-2mm} Comparisons on UCF-Crime Dataset \cite{ucf}}
    
    \resizebox{0.9\linewidth}{!}{
    
    \begin{tabular}{c|l c c| c}
        \textbf{Sup.}& \textbf{Method} & \textbf{Venue} & \textbf{Feature} & \textbf{AUC@ROC} $\uparrow$\\
        \toprule
        \multirow{5}{*}{\rotatebox{0}{Un-}} 
        & Lu et al.~\cite{lu2013abnormal}            & ICCV`13   & C3D       & 65.51\\ 
        & Hasan \cite{hasan}                       & CVPR`16   & -         &  50.60\\
        & BODS~\cite{wang2019gods}         & ICCV`19   & I3D       & 68.26\\
        & GODS~\cite{wang2019gods}         & ICCV`19   & I3D       & 70.46\\
        & GCL~\cite{zaheer2022generative}  & CVPR`22   & ResNext   &  71.04\\
        \hline
        \multirow{1}{*}{\rotatebox{0}{Fully-}} & \multirow{1}{*}{Liu \& Ma~\cite{liu2019exploring}} &  \multirow{1}{*}{MM`19}  & \multirow{1}{*}{NLN} &  \multirow{1}{*}{82.0}\\ 
        \hline
        \multirow{14}{*}{\rotatebox{0}{Weakly-}}
        &   GCN~\cite{gcn}                         & CVPR`19   & TSN       & 82.12  \\
        &   GCL$_{WS}$~\cite{zaheer2022generative} & CVPR`21   & ResNext   & 79.84 \\
        & Purwanto et al.  \cite{Purwanto_2021_ICCV}              & ICCV`21   & TRN       & \textul{85.00}\\
        &  Thakare et al. \cite{THAKARE2022117030}               & ExpSys`22 & C3D+I3D   & 84.48 \\
            \cline{2-5}
        & Sultani et al. \cite{ucf}                             & CVPR`18   & \multirow{5}{*}{\rotatebox{0}{C3D}}       & 75.41\\
        & Zhang et al.  \cite{zhang2019temporal}               & ICIP`19   &        & 78.70 \\
        &   GCN~\cite{gcn}                         & CVPR`19   &        & 81.08  \\
        &   CLAWS~\cite{claws}                     & ECCV`20   &        & 83.03 \\
        &   RTFM~\cite{rtfm}                     & ICCV`21   &        & 83.28 \\
            \cline{2-5}
        & Sultani et al.  \cite{ucf}                             & CVPR`18   &
        \multirow{6}{*}{\rotatebox{0}{I3D}}       & 77.92\\
        & Wu et al.  \cite{xd}                              & ECCV`20   &        & 82.44 \\
        &   DAM~\cite{dam}                         & AVSS`21   &        & 82.67 \\
        &   RTFM~\cite{rtfm}                       & ICCV`21   &        & 84.30 \\
        &  Wu \& Liu \cite{wu}                              & TIP`21    &        & 84.89  \\
        &   BN-SVP~\cite{Sapkota_2022_CVPR}        & CVPR`22   &        & 83.39 \\
            \cline{2-5}
        &   \multicolumn{2}{c }{\textbf{Ours: CLIP-TSA}} & CLIP & \textbf{87.58}\\ \bottomrule
    \end{tabular}
    \label{tb:UCF}
}
\end{table}
\vspace{-2mm}

\begin{table}[tb]
    \centering
   \caption{\vspace{-2mm} Comparisons on XD-Violence Dataset~\cite{xd}} 
    \resizebox{\linewidth}{!}{
    \begin{tabular}{c|c|l c c| c}
        \textbf{Sup.}& \textbf{Modality}& \textbf{Method} & \textbf{Venue} & \textbf{Feature} & \textbf{AUC@PR} $\uparrow$\\
        \toprule
        \multirow{2}{*}{\rotatebox{0}{Un-}} & --
        & OCSVM~\cite{ocsvm}                                  & NeurIPS`00    &  -- &   27.25 \\
        & & Hasan et al.~\cite{hasan}                   & CVPR`16       & --  & 30.77 \\
        \hline
         \multirow{9}{*}{\rotatebox{0}{Weakly-}} 
         & \multirow{5}{*}{\rotatebox{0}{\shortstack{Vision \\\& Audio}}} & Wu et al.~\cite{xd}                    & ECCV`20       & I3D(V) + VGGish(A) & 78.64 \\
        & & Wu \& Liu~\cite{wu}                    & TIP`21        & I3D(V) + VGGish(A) & 75.90 \\
        & & Pang et al.~\cite{pang2021icassp}        & ICASSP`21     & I3D(V) + VGGish(A) & 81.69 \\
        & & MACIL-SD~\cite{yu2022macil}  & MM`22         & I3D(V) + VGGish(A) & \textul{83.40} \\
        & & DDL~\cite{cmala}             & ICCECE`22     & I3D(V) + VGGish(A) & \textbf{83.54} \\ \cline{2-6}\cline{2-6}
        & \multirow{4}{*}{\rotatebox{0}{Vision}} 
        & Sultani et al.~\cite{ucf}                   & CVPR`18       & C3D(V)& 73.20 \\
        & & RTFM~\cite{rtfm}             & ICCV`21       & C3D(V)& 75.89 \\
        & & RTFM~\cite{rtfm}             & ICCV`21       & I3D(V) & 77.81 \\\cline{3-6}
      & & \multicolumn{2}{c }{\textbf{Ours: CLIP-TSA}} & CLIP(V) & {82.19}  \\ \bottomrule
    \end{tabular}}
    \label{tb:XD}
\end{table}

\begin{table}[tb]
    \centering
    \caption{Comparisons on ShanghaiTech Campus Dataset~\cite{shanghaitech}}
    \resizebox{0.9\linewidth}{!}{
    \begin{tabular}{c|l c c | c}
        \textbf{Sup.} & \textbf{Method} & \textbf{Venue} & \textbf{Feature} & \textbf{AUC@ROC} $\uparrow$\\
        \toprule
        \multirow{4}{*}{\rotatebox{90}{Un-}} & 
        Hasan et al.~\cite{hasan}                               & CVPR`16   & - & 60.85 \\
        & Gao et al.~\cite{gao2021memory}                     & ICCV`19   & - & 71.20 \\
        & Yu et al.~\cite{yu2020cloze}                       & MM`20     & - & 74.48\\
        & GCL$_{PT}$~\cite{zaheer2022generative}     & CVPR`21   & ResNext & 78.93 \\
        \hline
        \multirow{7}{*}{\rotatebox{90}{Weakly-}}
        & GCN~\cite{gcn}                           & CVPR`19 & TSN      & 84.44 \\
        & GCL$_{WS}$~\cite{zaheer2022generative}   & CVPR`21 & ResNext  & 86.21 \\
        & Purwanto et al.~\cite{Purwanto_2021_ICCV}                & ICCV`21 & TRN      & 96.85 \\
        \cline{2-5}
        & GCN~\cite{gcn}                           & CVPR`19 & \multirow{5}{*}{\rotatebox{0}{C3D}} & 76.44 \\
        & Zhang et al.~\cite{zhang2019temporal}                 & ICIP`19 &       & 82.50 \\
        & CLAWS~\cite{claws}                       & ECCV`20 &       & 89.67 \\
        & RTFM~\cite{rtfm}                         & ICCV`21 &       & 91.57 \\
        & BN-SVP~\cite{Sapkota_2022_CVPR}          & CVPR`22 &       & 96.00 \\
        \cline{2-5}
        & Sultani et al.~\cite{ucf}                              & CVPR`18 &       & 85.33 \\
        & AR-Net~\cite{9102722}                    & ICME`20 &  \multirow{3}{*}{\rotatebox{0}{I3D}}     & 91.24 \\
        & DAM~\cite{dam}                           & AVSS`21 &       & 88.22 \\
        & Wu \& Liu~\cite{wu}                                & TIP`21  &       & \textul{97.48} \\
        & RTFM~\cite{rtfm}                         & ICCV`21 &       & 97.21 \\
        
        \cline{2-5}
        &  \multicolumn{2}{c}{\textbf{Ours: CLIP-TSA}} & CLIP & \textbf{98.32} \\ \bottomrule
    \end{tabular}
    }
    \label{tb:SH}
\end{table}



\section{Experimental Results}
\noindent
\hspace{5mm}\textbf{A. Dataset and Implementation Details}
\label{sec:impdet}
We conduct the experiment on three datasets, \textit{i.e.}, UCF-Crime \cite{ucf}, ShanghaiTech \cite{shanghaitech}, XD-Violence \cite{xd}, as summarized in Table~\ref{tb:datasets}.

Following \cite{ucf}, we divide each video into 32 video snippets, \textit{i.e.,} $T = 32$ (Eq.~\ref{eq:reshape}) and snippet length $\delta = 16$. The scorer network $\theta_s$ (Sec.~\ref{sec:tsa}) is defined as an MLP of three layers (512, 256, 1). Leveraging CLIP \cite{radford2021learning}, $d$ is set as 512. We set $M = 100$ for Gaussian noise in Eq.~\ref{eq:gau}.
Let $\kappa = \lfloor T \times r \rfloor$; $r$ is a hyperparameter for which we choose 0.7. 
Our CLIP-TSA is trained in an end-to-end manner using PyTorch. We use Adam optimizer \cite{adam} with a learning rate of 0.001, a weight decay of 0.005, and a batch size of 16 for all datasets.
\textbf{B. Performance Comparison}
Tables \ref{tb:UCF}, \ref{tb:XD}, and \ref{tb:SH} show the performance comparison between our proposed method and other SOTAs on multiple datasets. In all tables, the best scores are \textbf{bolded}, and the runner-ups are \textul{underlined}. 
Our remarkable performance mainly comes from the two modules of: (i) rich contextual vision-language feature CLIP and (ii) TSA. To analyze the impact of each module as well as make a fair comparison, we conduct a comprehensive ablation study as in Table \ref{tb:ablation}. In this table, we benchmark each module by: (i) replacing the CLIP feature with C3D~\cite{C3D} and I3D~\cite{i3d_2017} and (ii) switching on and off our TSA module. By comparing Table \ref{tb:ablation} against Tables \ref{tb:UCF}-\ref{tb:SH}, we can see that C3D-TSA (replace CLIP feature with C3D) and I3D-TSA (replace CLIP feature with I3D) still outperform the other SOTA models benchmarked with C3D and I3D features on UCF-Crime and ShanghaiTech datasets. As for XD-Violence dataset, we outperform all other vision-based (V) methods, and the SOTAs with a higher score commonly leverage extra modality, audio (A), using VGGish. Evidently, Table~\ref{tb:ablation} shows: (i) performance improvement for all scenarios with our TSA on, and (ii) the best performance is obtained by the proposed CLIP-TSA thanks to the contribution of both strong CLIP feature and TSA module.

\begin{table}
    \centering
    \caption{Ablation Study of TSA on Various Features}
    \resizebox{0.85\linewidth}{!}{
    \begin{tabular}{c | c || c | c | c}
        \multirow{2}{*}{\textbf{Feature}} &
        \multirow{2}{*}{\textbf{TSA}} & 
        \textbf{UCF-Crime} & 
        \textbf{ShanghaiTech} &
        \textbf{XD-Violence} \\ && 
        (AUC@ROC $\uparrow$) & 
        (AUC@ROC $\uparrow$) & 
        (AUC@PR $\uparrow$) \\
        
        \toprule
        C3D     & \xmark    &  82.59                    &  96.73                    &  76.84 \\
        C3D     & \cmark    &  {83.94}           &  {97.19}           &  {77.66} \\
        \hline
        I3D     & \xmark    &  83.25                    &  96.39                    &  77.74 \\
        I3D    & \cmark    &  {84.66}           &  {97.98}           &  {78.19} \\
        \hline
        CLIP     & \xmark    &  \textul{86.29}           & \textul{98.18}            &  \textul{80.43}\\
        CLIP     & \cmark    & {\textbf{87.58}}  & {\textbf{98.32}}   &  {\textbf{82.19}}\\
        \bottomrule
    \end{tabular}
    }
    \label{tb:ablation}
\end{table}

\noindent

\textbf{Conclusion}
The paper presents CLIP-TSA, an effective end-to-end weakly-supervised VAD framework. Specifically, we proposed the novel TSA mechanism that maximizes attention on a subset of features while minimizing attention on noise and showed its applicability to this domain. We also applied TSA to CLIP-extracted features to demonstrate its efficacy in Vision-Language features and exploited it in the problem.  Comprehensive experiments and ablation studies empirically validate the excellence of our model on the three popular VAD datasets by comparing ours against the SOTAs.

\textbf{Acknowledgements}
This paper is supported by the National Science Foundation under Award No OIA-1946391 RII Track-1, NSF 1920920 RII Track 2 FEC, NSF 2223793 EFRI BRAID, NSF 2119691 AI SUSTEIN, and NSF 2236302.


\scriptsize
\bibliographystyle{IEEEbib}
\bibliography{strings,refs}

\end{document}